\documentclass[letterpaper, 10 pt, conference]{ieeeconf}  

\IEEEoverridecommandlockouts                              

\usepackage{graphicx}
\usepackage{hyperref}
\usepackage{color,soul}
\usepackage[usenames,dvipsnames]{xcolor}
\usepackage{caption}
\usepackage{subcaption}
\usepackage{multirow}
\usepackage{array}
\usepackage{booktabs}
\usepackage{graphicx}
\usepackage{caption}
\usepackage{tabularx}
\usepackage{float}
\usepackage[normalem]{ulem}
\usepackage{cuted}
\usepackage{fancyhdr}

\newcommand{\ieeecopyright}{%
  \footnotesize
  \centering
  \copyright~2025 IEEE.  Personal use of this material is permitted.  Permission from IEEE must be obtained for all other uses, in any current or future media, including reprinting/republishing this material for advertising or promotional purposes, creating new collective works, for resale or redistribution to servers or lists, or reuse of any copyrighted component of this work in other works.
}

\fancypagestyle{ieeefirstpage}{%
  \fancyhf{}
  \fancyfoot[C]{\ieeecopyright}
}

\hypersetup{
    colorlinks=true,        
    linkcolor=black,         
    citecolor=blue,         
    urlcolor=blue           
}

\newcommand{\kimia}[1]{\textcolor{black}{#1}}
\newcommand{\kw}[1]{\textcolor{black}{#1}}
\newcommand{\parsa}[1]{\textcolor{black}{#1}}
\newcommand{\parsaa}[1]{\textcolor{black}{#1}}

\newcommand{\skw}[1]{\textcolor{black}{#1}}
\newcommand{\SYS}[1]{\textcolor{black} {FastTrack}}

\newcommand{\InjectComment}[1]{\textcolor{black}{#1}}

\graphicspath{{./figures/}}

\title{\LARGE \bf
\SYS{}: GPU-Accelerated Tracking for Visual SLAM
}

\author{Kimia Khabiri$^{1}$, Parsa Hosseininejad$^{1}$, Shishir Gopinath$^{1}$, Karthik Dantu$^{2}$, Steven Y. Ko$^{1}$ \\
\thanks{$^{1}$Simon Fraser University. 
\textcolor{black}\texttt{\{\href{mailto:kka156@sfu.ca}{\textcolor{black}{kka156}}, \href{mailto:sph6@sfu.ca}{\textcolor{black}{sph6}}, \href{mailto:sgopinat@sfu.ca}{\textcolor{black}{sgopinat}}, \href{mailto:steveyko@sfu.ca}{\textcolor{black}{steveyko}}\}@sfu.ca}} 
\thanks{$^{2}$University at Buffalo. 
\texttt{\href{mailto:kdantu@buffalo.edu}{\textcolor{black}{kdantu@buffalo.edu}}}}
}

\begin{document}

\maketitle
\thispagestyle{ieeefirstpage}
\pagestyle{empty}

\begin{abstract}


The tracking module of a visual-inertial SLAM system processes incoming image frames and IMU data to estimate the position of the frame in relation to the map. It is important for the tracking to complete in a timely manner for each frame to avoid poor localization or tracking loss. We therefore present a new approach which leverages GPU computing power to accelerate time-consuming components of tracking in order to improve its performance. These components include stereo feature matching and local map tracking. We implement our design inside the ORB-SLAM3 tracking process using CUDA. Our evaluation demonstrates an overall improvement in tracking performance of up to 2.8$\times$ \parsaa{on a desktop and Jetson Xavier NX board} in stereo-inertial mode, using the well-known SLAM datasets EuRoC and TUM-VI.
\end{abstract}

\section{Introduction}


Visual-inertial simultaneous localization and mapping (SLAM) systems estimate the pose of a robot or device and the locations of features in its surrounding environment using mono, stereo, or RGB-D images and inertial measurement unit (IMU) data. Visual SLAM is used as a service in diverse applications including autonomous robots as well as augmented and virtual reality devices. Tracking refers to the process of extracting features from incoming frames, preintegrating IMU data, finding correspondences between the current frame, previous frame, and the map, and using this information to estimate the current pose of the system.

In SLAM systems, since sensor data must first flow through tracking before information can be provided to other modules or threads, poor tracking performance serves as a bottleneck for the entire SLAM system. In real-time scenarios, this can also lead to dropped frames resulting in inaccurate localization, loop detection failure, or, in more extreme cases, tracking loss~\cite{Semenova2024}, especially on resource-constrained devices. Thus, it is important for tracking to run as quickly as possible in order to keep up with changes to the state of the system and the environment. It is also desirable for tracking to be consistent in processing time so that the rest of the system works in a timely manner. 


To address this challenge, we introduce \SYS{}, a visual-inertial tracking module that leverages GPU computing power to accelerate the most time-consuming steps of tracking. In doing so,
\skw{we carefully examine all algorithms and components used for tracking, analyze their individual performance impacts, and analyze the feasibility of offloading their computation to the GPU. We then devise several techniques to efficiently run them on the GPU and demonstrate the benefits of our techniques using well-known datasets.
The conventional criterion for offloading a computation to the GPU \kimia{is how well it can be parallelized}. However, it is \kimia{also} necessary to consider \kimia{the cost of data transfer between the CPU and GPU as the time spent transferring data may offset the performance gains achieved through parallel processing}. Thus, we carefully pick and choose which components to run on the CPU or the GPU based on their parallelizability and data transfer costs.}



Our design combines several GPU tasks (kernels) that offload different parts of the tracking process to the GPU. This includes (i) kernels to accelerate stereo feature matching for pinhole and fisheye cameras, and (ii) kernels to offload the task of \kimia{searching local map points in the current frame}. 
To further improve performance, we efficiently handle data transfers between tracking components, bypass the pose optimization task in local map tracking, and integrate an
existing acceleration technique for ORB feature extraction~\cite{Muzzini2023}.


\skw{We base our implementation on ORB-SLAM3 and compare the performance of our acceleration techniques using EuRoC~\cite{euroc} and TUM-VI~\cite{tumvi} on a desktop and an embedded platform.}
Our results show that our implementation is \parsaa{up to} $2.8\times$ faster in stereo-inertial mode, while producing comparable trajectory errors. \parsa{Our code is available at \url{https://github.com/sfu-rsl/FastTrack}.}


Though our implementation is based on ORB-SLAM3 \cite{orbslam3}, we expect our ideas to
translate to a broader class of SLAM systems, sometimes referred to as feature-based SLAM
systems. This class of systems represents one of the several SLAM architectures that exist
today~\cite{macario2022comprehensive}, and utilizes classic feature detection and multi-view
geometry along with bundle adjustment and loop closure. Examples of such systems include
ORB-SLAM2~\cite{orbslam2}, ORB-SLAM3~\cite{orbslam3}, Kimera~\cite{rosinol2021kimera},
OKVIS~\cite{leutenegger2015keyframe}, and numerous derivatives of these systems developed over the years.

\section{Background} \label{sec:background}

\begin{figure}[t]
    \includegraphics[width=1\linewidth]{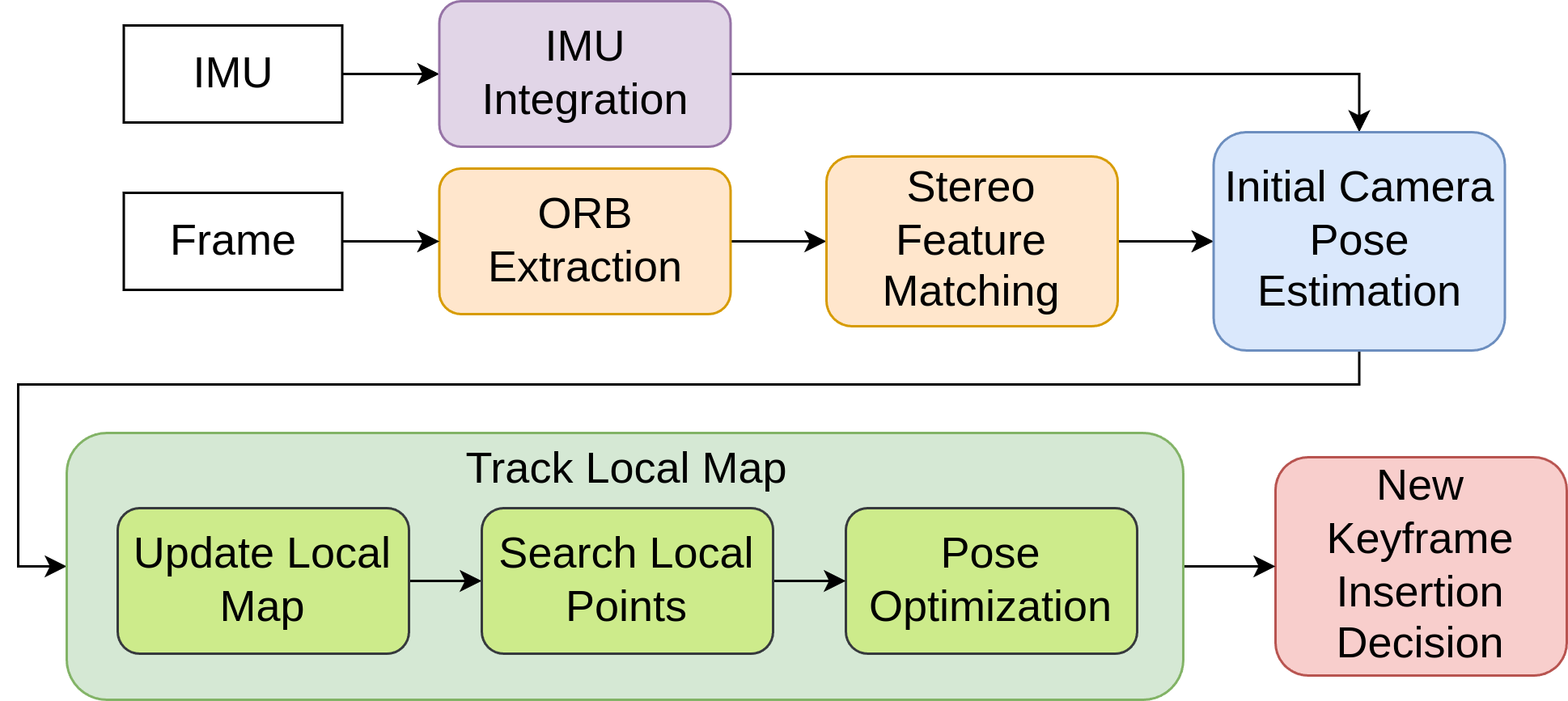}
    \centering
    \caption{The workflow of the tracking process in ORB-SLAM3.}
    \label{fig:tracking_thread_components}
\end{figure}

To establish the context of our techniques, we first provide a brief overview of how tracking works and introduce its components.
Tracking is responsible for extracting features from incoming frames, relating these features to the previous frame as well as the map, and estimating the current pose of the system. \autoref{fig:tracking_thread_components} provides a visual representation of tracking and its components. In what follows, we will introduce the components essential for this paper. \parsa{In \SYS{}, we offload all or parts of these components to the GPU.}

\subsection{ORB Extraction}
During this phase, the system generates multiple scales of the same image, known as the image pyramid, and then extracts features\kimia{, referred to as keypoints,} from all of the different scales. Each keypoint is then described by a corresponding descriptor, which encodes its unique characteristics. The system then filters out keypoints with less information.

\subsection{Stereo Matching}
\label{subsec:stereo_match}


The goal of this component is to match each keypoint from the left image with the most similar \kimia{keypoint} in the right image to calculate depth. 
Generally, stereo matching employs different strategies based on the camera type. ORB-SLAM3 handles two camera types: pinhole and fisheye.



\kimia{
In the case of fisheye cameras, the matching process involves a brute-force search. For each keypoint identified in the left image, the algorithm iterates through all the keypoints in the right image to find the best match.
}

\kimia{
For pinhole cameras, the search is more efficient because the matched keypoints are expected to lie along the epipolar lines. Instead of searching through all points in the right image, the algorithm restricts the search to a specific region along these lines. This reduces the search space, making stereo matching faster and more computationally efficient compared to the brute-force approach in fisheye cameras.
}

\subsection{Initial Camera Pose Estimation}
Initial Pose Estimation aims to predict the camera's pose based on the previous frame's position. To achieve this, the system conducts a process called Search by Projection, in which it projects the previous frame's map points into the current frame and searches for correspondences between these map points and the current frame's features. The system then refines the current frame's pose based on the identified correspondences.

\subsection{Track Local Map} 
\label{subsec:tracK_local_map}
\kimia{The Track Local Map component optimizes the camera's position in relation to a local map. The local map consists of a subset of map points—3D features in the environment—that are relevant to the current camera pose. Track Local Map consists of three stages: Update Local Map, Search Local Points, and pose optimization.}


In the Update Local Map stage, local map points near the current frame are collected by first gathering local keyframes that observe the frame's map points, and then retrieving the map points observed by those keyframes.

Within the Search Local Points process, the key task is Search by Projection. This task involves projecting local map points into the current frame and finding correspondences between the projected map points and the features in the frame.

Finally, Pose Optimization refines the camera pose and associated inertial parameters using all the map points found in the frame.


\section{Related Work}

There have been several efforts to accelerate the components of visual SLAM
using the on-device GPU. A major area of interest is feature tracking.
Aldegheri et al.~\cite{Aldegheri2019} develop a graph representation of
ORB-SLAM2~\cite{orbslam2} by observing dataflow, and use this to efficiently
subdivide computations between the CPU and GPU, allowing for real-time
performance on an NVIDIA Jetson TX2 embedded board. Muzzini et
al.~\cite{Muzzini2023} accelerate the ORB extraction step in tracking,
and propose a method for image pyramid construction, where each level is
simultaneously computed by the same kernel.  Gopinath et
al.~\cite{Gopinath2023} offload computationally intensive operations in
local bundle adjustment, and Kumar et al.~\cite{Kumar2024} offload pose
graph optimization to the GPU. 

Kumar et al.~\cite{KumarPark24} propose Jetson-SLAM, a frontend-middle-end
SLAM system design that accelerates tracking, including stereo-matching and
conic projection, using the GPU. Their technique is based on ORB-SLAM2, which
lacks the advanced features of ORB-SLAM3 such as support for fisheye
camera, combined visual-inertial information, and multiple maps. Since our
technique is based on ORB-SLAM3, we support all those features in our GPU
acceleration. In addition, we carefully examine the data transfer costs
and dependencies across all the components that provide all the advanced
features, which results in a significantly different design from that of
Kumar et al. We provide a detailed performance comparison between the two
systems in \autoref{subsec:comparison}, where \SYS{} outperforms Jetson-SLAM
by achieving a higher FPS with consistently lower trajectory error. Even
when Jetson-SLAM delivers higher FPS in some sequences, its increased
trajectory error demonstrates \SYS{}’s superior balance of speed and
accuracy in real-time stereo SLAM.

\section{Design} \label{sec:design}

\begin{figure}[t] 
    \centering
    \begin{subfigure}{0.75\linewidth} 
        \includegraphics[width=\textwidth]{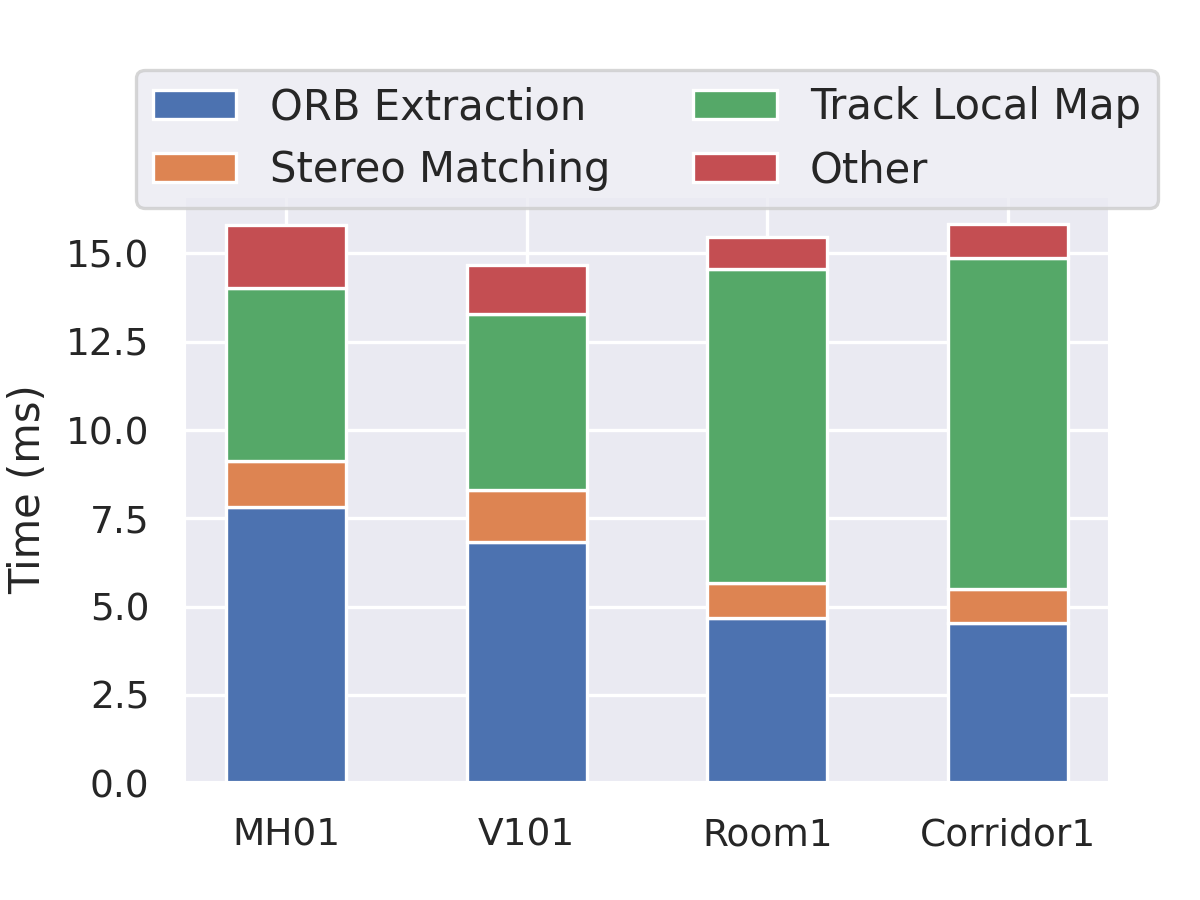}
        \caption{Tracking}   
        \label{fig:tracking_breakdown}
    \end{subfigure}%
    \\
    \begin{subfigure}{0.75\linewidth} 
        \includegraphics[width=\textwidth]{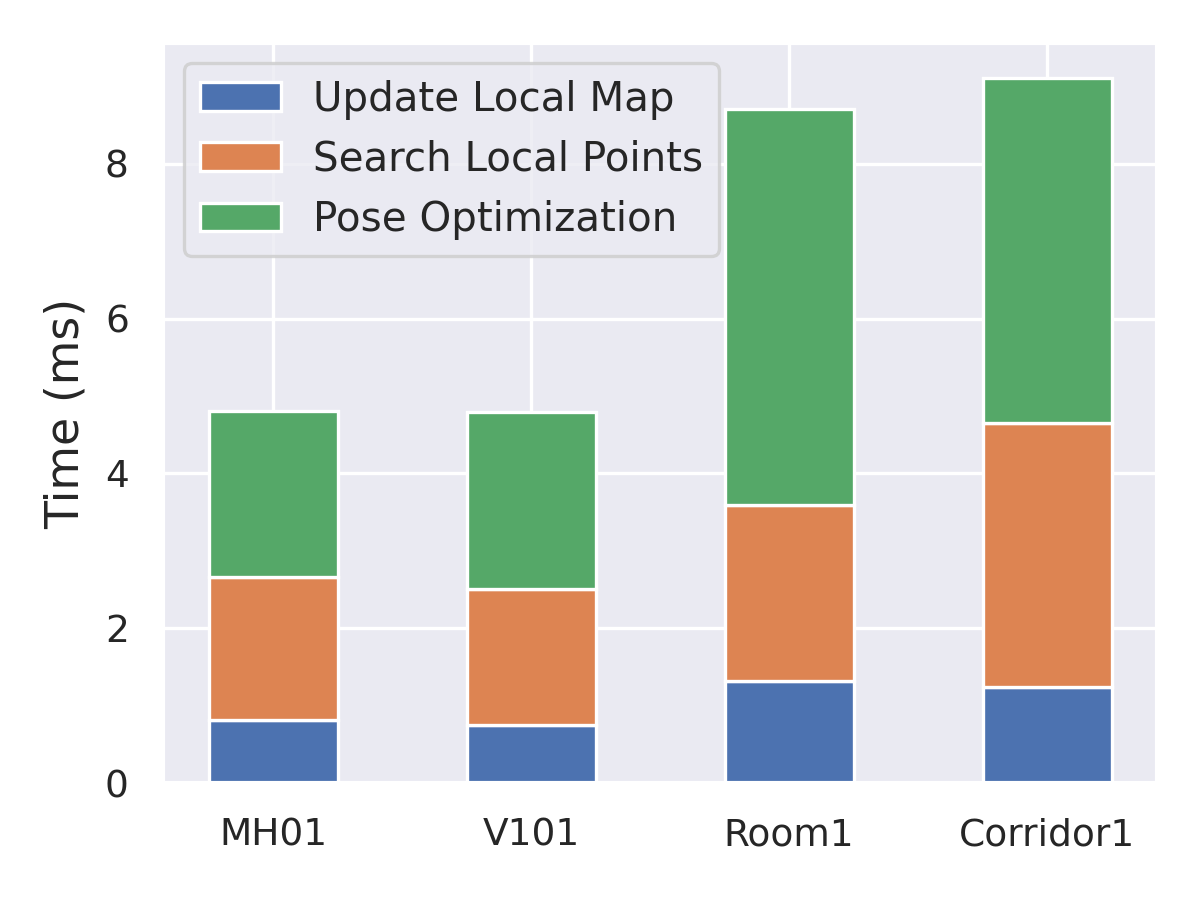}
        \caption{Track Local Map}
        \label{fig:track_local_map_breakdown}
    \end{subfigure}
    \caption{Time spent in different components of tracking.}
    \label{fig:tracking_time_breakdown}
\end{figure}

\begin{figure*}[t]
    \centering
    \includegraphics[width=.9\textwidth]{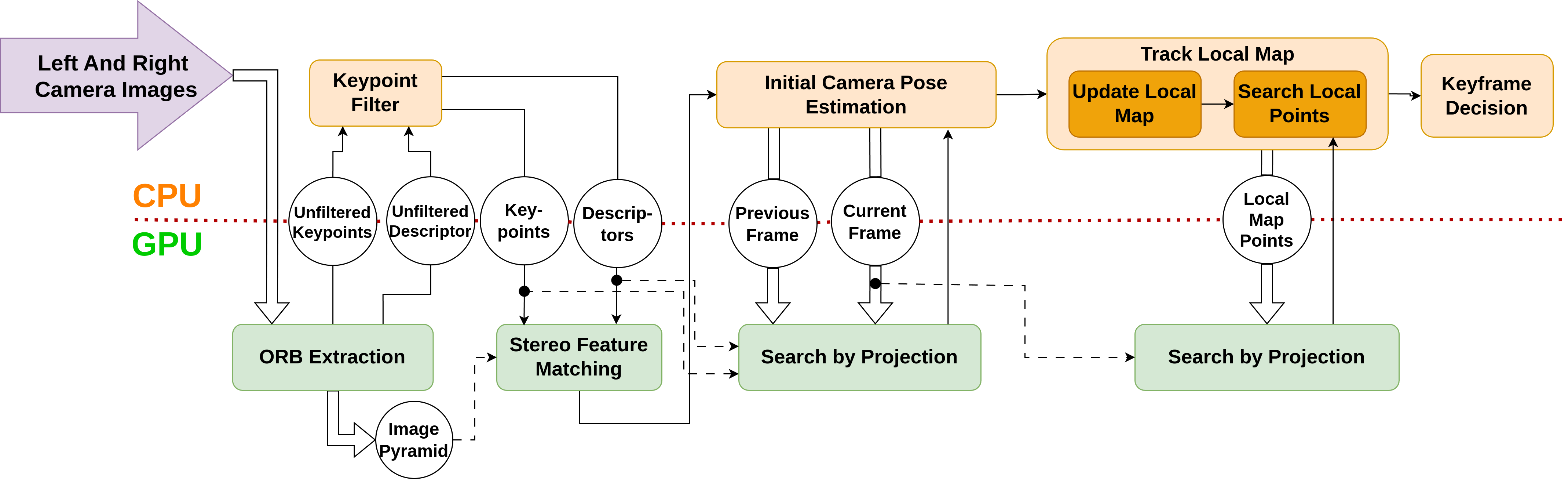}
    \caption{Data flow of the tracking process in \SYS{}: CPU components are at the top, GPU components at the bottom. Data transfers are represented using three distinct types of arrows: \kw{dotted arrows represent no data transfer, regular arrows represent lightweight data transfer, and thick arrows represent heavy data transfer.}
    }
    \label{fig:gpu_tracking_thread_dataflow}
\end{figure*}

To accelerate the tracking process, we aim to reduce the execution time of its resource-intensive components. As shown in \autoref{fig:tracking_breakdown}, tracking the local map, ORB extraction, and stereo matching are the three most time-consuming components inside tracking. In the EuRoC dataset, ORB extraction consumes the majority of the tracking time, whereas in the TUM-VI dataset, the most demanding task is tracking the local map, which takes twice as long as in EuRoC. This difference stems from the distinct algorithms for handling different camera types used in the EuRoC and TUM-VI datasets, pinhole and fisheye, respectively.

\subsection{Overview}
To enhance the performance of the tracking process, we offload the stereo matching and Track Local Map components to the GPU and reuse an existing GPU implementation of ORB extraction \cite{Muzzini2023}.
Within the Track Local Map component, we offload the Search Local Points task to the GPU and disable pose optimization. We select these components for optimization as they are the most time-consuming tasks within tracking the local map process, as shown in \autoref{fig:track_local_map_breakdown}.

Our decision to offload components to the GPU is driven by their potential for performance improvement through parallelization, while ensuring minimal overhead \skw{for CPU-to-GPU data transfer or vice versa.} Note that each component offloaded to the GPU enables further offloading opportunities by reducing data transfer costs for subsequent components. \autoref{fig:gpu_tracking_thread_dataflow} provides an overview of tracking's data flow in \SYS{}. \parsa{
In the figure, boxes represent components running on either the CPU (top) or the GPU (bottom). Circles denote the data generated by each component, while arrows indicate data transfers between them. \kw{Regular and thick arrows represent lightweight and heavyweight data transfers, respectively, while dotted arrows indicate no data transfer.} We refer to this figure throughout this section and discuss its details. }

In all of our kernels, to minimize overhead, we eliminate redundant GPU memory allocations by allocating the required memory for various data at the start of the system's execution. We reuse this memory throughout the execution and release it before the system shuts down.

\subsection{ORB Extraction}
\skw{ORB Extraction is the first component that processes camera input as shown in \autoref{fig:gpu_tracking_thread_dataflow}.}
In \SYS{}, ORB Extraction runs on the GPU. There are two main reasons for this. First, a well-optimized GPU implementation~\cite{Muzzini2023} already exists, which we can reuse. Second, by generating the image pyramid on the GPU, ORB Extraction reduces the data transfer cost for stereo matching, which relies on this data as input.
The image pyramid consists of multiple scales of the same image. \kimia{This size is substantial for the stereo matching kernel, as it requires the image pyramids for both the left and right images, which together comprise 90\% of the total data required.}
As illustrated in \autoref{fig:gpu_tracking_thread_dataflow}, the ORB Extraction kernel produces image pyramid data and stores it in the GPU memory, \skw{which we later reuse for stereo matching.} \kimia{By doing so, we reduce the data transfer costs for the stereo matching kernel massively, reducing it from 0.80 ms to 0.08 ms on our desktop machine.}
Keypoints and descriptors are also produced by ORB Extraction and used by stereo matching. However, the aforementioned ORB Extraction kernel we reuse~\cite{Muzzini2023} does not offload the keypoint filtering stage to the GPU since the computation is not suitable for parallelization. Thus, we transfer keypoints and descriptors from the CPU memory for stereo matching.

\subsection{Stereo Matching}
\label{design_stereo_match}
Stereo matching is the second component shown in \autoref{fig:gpu_tracking_thread_dataflow} and \SYS{} offloads it to the GPU. The primary goal of stereo matching is to identify a corresponding keypoint in the right image for each keypoint in the left image. This process is inherently parallelizable, as each keypoint can be matched independently, with no data dependencies between matches. In addition, a significant portion of the required data is already in GPU memory as explained earlier.

Stereo matching handles pinhole and fisheye cameras differently and our GPU design handles them separately as well.
For pinhole, we break down its functionality into two separate kernels: one for identifying the best match for each keypoint, and the other for refining this match. This separation allows us to customize the design for each kernel, \parsaa{ which in turn enables us to optimize the number of GPU threads and utilize shared memory when needed to improve the overall performance}. 
In the first kernel, we launch \kimia{one thread per keypoint in the left image}, with each thread tasked to find the best keypoint match for a \emph{single} keypoint in the left image within the right image.
\parsaa{In the second kernel, we launch \kimia{many blocks} of threads, each block tasked to refine a single match identified \kimia{in the first kernel}. We use shared memory in each block to refine the match and find the best candidate among the nearby pixels. This design confines each GPU thread to performing a few simple arithmetic operations, significantly boosting performance. Moreover, by leveraging shared memory—which has approximately 100× lower latency than global memory \cite{nvidia_shared_memory}—we minimize memory access times and enable efficient data sharing among threads.}

For the fisheye setup, there is no need for two kernels since no refinement phase is required. Thus, we use a single kernel, similar to the first kernel in the pinhole version.

\subsection{Update Local Map}
Update Local Map is the first component within Track Local Map as shown in \autoref{fig:gpu_tracking_thread_dataflow}, and \SYS{} keeps this component on the CPU. This is because, although the process involves large loops, the execution cost per iteration is low,
while GPU offloading causes a large amount of data transfer.

As explained in \autoref{sec:background}, Update Local Map has two steps. In the first step, it iterates over the current frame’s map points, \parsa{collects keyframes which observe these map points, and copies them} into a set of local keyframes.
The execution cost is minimal as it only has copy operations. However,
we need to transfer \kimia{all of the keyframes that observe any of the frame's map points} \kimia{in order to conduct this process}.

In the second step, the system collects map points observed by local keyframes into a set called local map points. Again, the execution cost is low as it only involves copy operations.
However, the data transfer is substantial \kimia{as we need to transfer all the map points observed by any of the local keyframes}.

Thus, we determine that the Update Local Map process is not well-suited for GPU optimization and keep it on CPU.

\subsection{Search Local Points}

Search Local Points is the last component in the tracking process shown in \autoref{fig:gpu_tracking_thread_dataflow}.
In \SYS{}, we offload \emph{Search by Projection}, a task within the Search Local Points component, to the GPU.
Search by Projection is suitable for GPU optimization because it has a high execution cost, which can be significantly reduced by GPU parallelization, while the data transfer costs associated with it are manageable.

Search by Projection is responsible for finding correspondences between local map points and frame features. It involves a series of costly iterations, with each iteration identifying a correspondence for a single map point. Since the map points are independent of each other, the result of one iteration does not affect others. This allows us to divide the task concurrently across multiple GPU threads.
 
As illustrated in \autoref{fig:gpu_tracking_thread_dataflow}, our system uses the Search by Projection kernel in two components: Search Local Points and \kimia{Initial} Pose Estimation.
This is because both components use the same Search by Projection algorithm and we can adapt our GPU kernel for both.
Search by Projection within \kimia{Initial} Pose Estimation deals with \InjectComment{heavy data transfer costs}, as it has to transfer the previous and current frames to the GPU. However, the Search by Projection process does not use much of the data within frame objects, allowing us to \kimia{reduce transfer costs by} only transfer\kimia{ring} necessary data.
\kimia{In addition, \parsaa {the} stereo matching kernel further reduces frame transfer costs by loading frame keypoints and descriptors data to the GPU in earlier stages, as illustrated in \autoref{fig:gpu_tracking_thread_dataflow}.} 

Search by Projection within Search Local Points also deals with a substantial amount of data, including the current frame and the local map points as inputs.
For the current frame, the transfer cost is zero, as the data is already on the GPU from the previous Search by Projection kernel.
For local map points, we organize the data more efficiently to eliminate unnecessary data transfers.
To do this, we break up the complex data structure that represents a map point into finer-grained structures. We then process those structures separately and transfer them as needed. 


\subsection{Pose Optimization}
\label{design_pose_optimization}
\begin{figure}[t]
    \centering
    \includegraphics[width=1\linewidth]{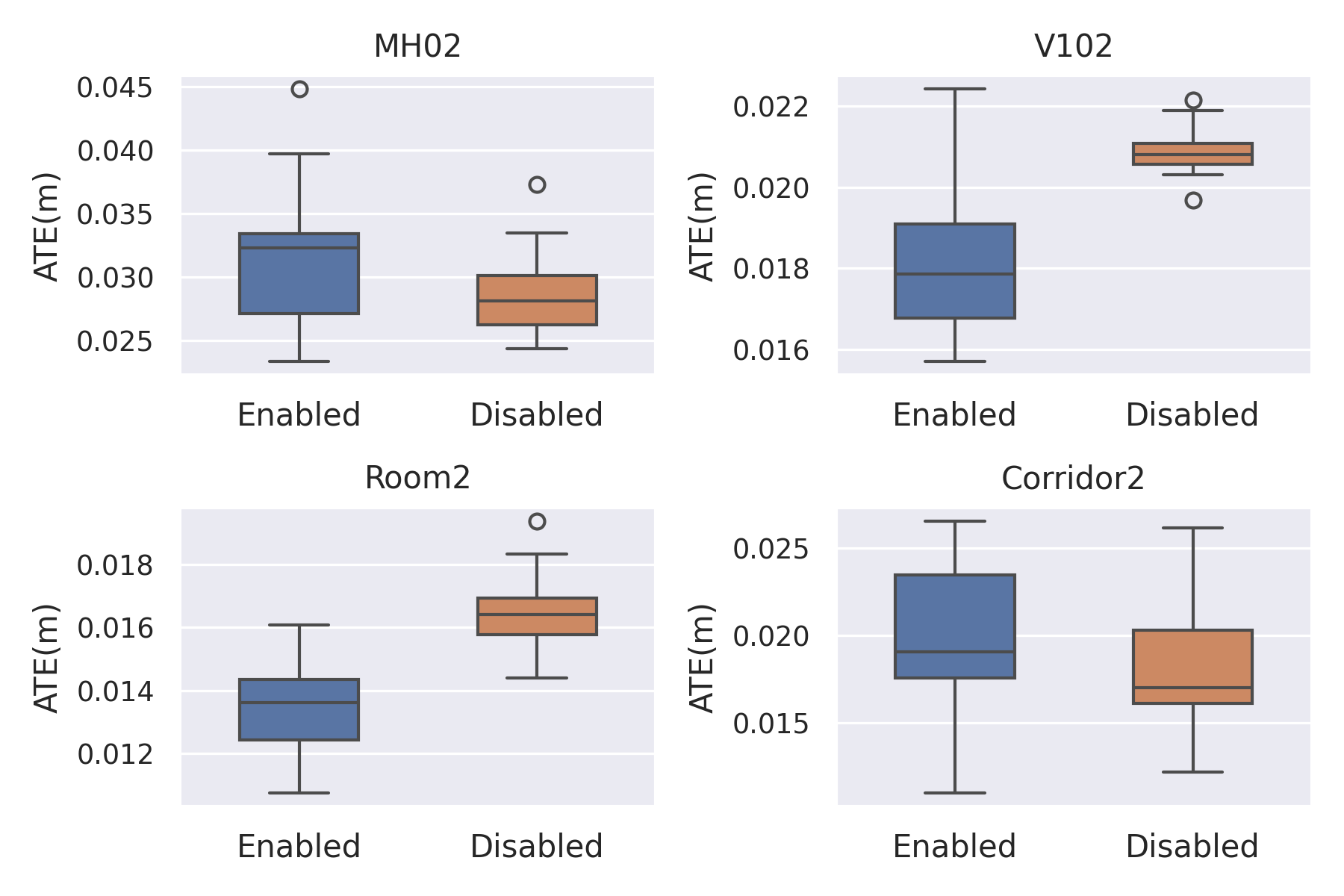}
    \caption{Absolute trajectory error comparison with Pose Optimization on and off in ORB-SLAM3.}
    \label{fig:ATE_pose_on_and_off}
\end{figure}

We do not offload Pose Optimization to the GPU, and instead we choose to disable it. As shown previously in \autoref{fig:track_local_map_breakdown}, Pose Optimization has the most overhead in the Track Local Map component. Disabling Pose Optimization greatly reduces the overall tracking time with minimal impact on the trajectory error. \autoref{fig:ATE_pose_on_and_off} shows a comparison of absolute trajectory error (ATE) with/without Pose Optimization in 20 runs for each sequence. In all cases, the differences in ATE are minimal, with only slight changes in error and variance. This suggests that disabling Pose Optimization has a negligible impact on ATE performance, possibly due to optimizations which occur further in the pipeline, such as local bundle adjustment. Therefore, we bypass this step when tracking the local map.

\section{Evaluation} \label{sec:evaluation}

\begin{table}[]
    \centering
    \small
    \begin{tabular}{c|cc}
         \bf Machine & \bf Specs\\
         \toprule 
         & 20-core Intel Core i7-12700K CPU @ 5.0 GHz\\
         Desktop & 10496-core NVIDIA RTX 3090 GPU\\
        & 64 GB RAM\\ 
        \midrule
        & 6-core ARM Carmel CPU @ 1.4 GHz 15W \\
        Xavier NX & 384-core NVIDIA Volta GPU \\
        & 8 GB RAM\\
         \bottomrule
    \end{tabular}
    \caption{Evaluation machine specifications.}
    \label{tab:specs}
\end{table}

\begin{figure}[t]
    \centering
    \includegraphics[width=.9\linewidth]{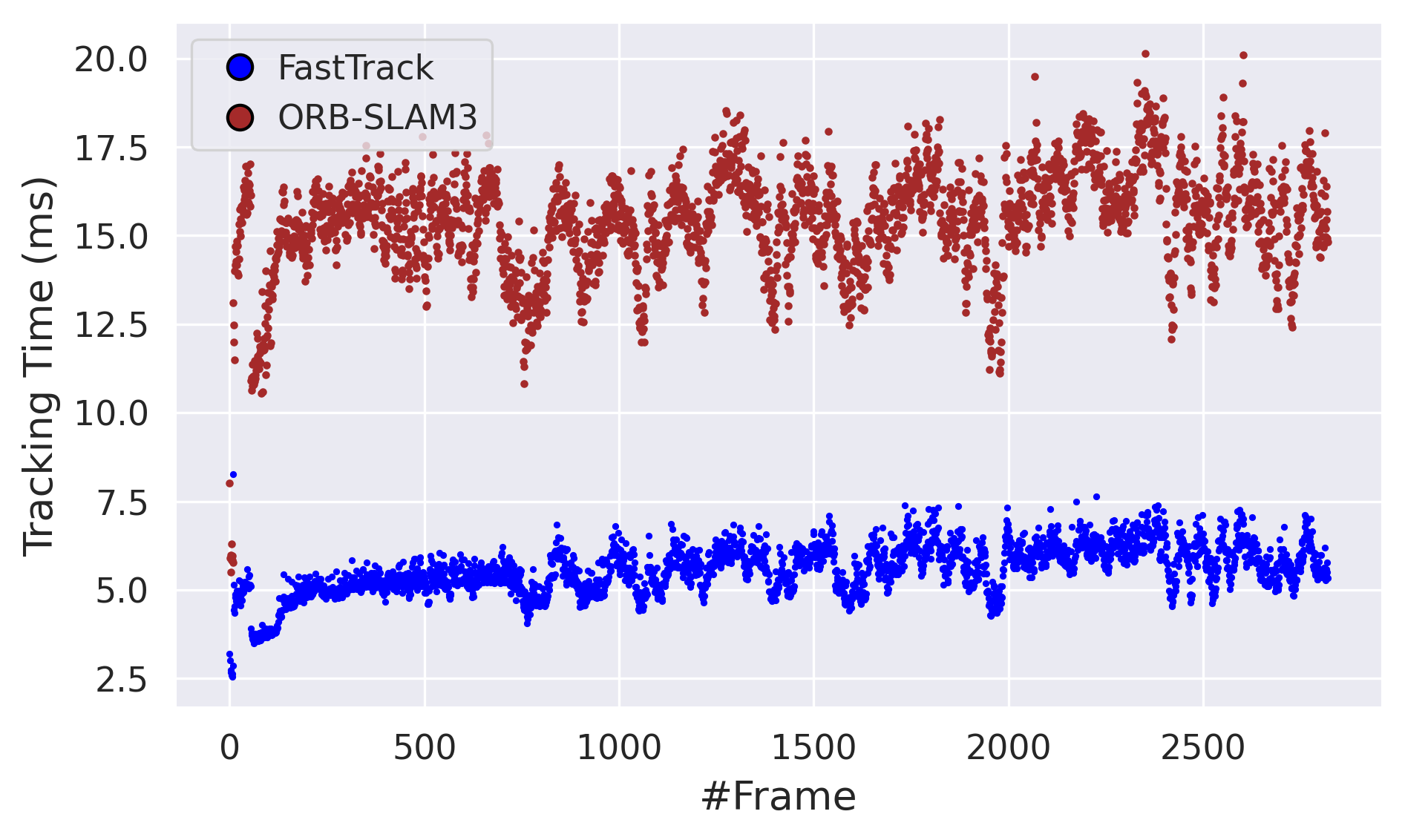}
    \caption{Room1 frame's tracking in the desktop setting.}
    \label{fig:room1_tracking_time}
\end{figure}

\begin{figure}[t]
    \centering
    \includegraphics[width=.9\linewidth]{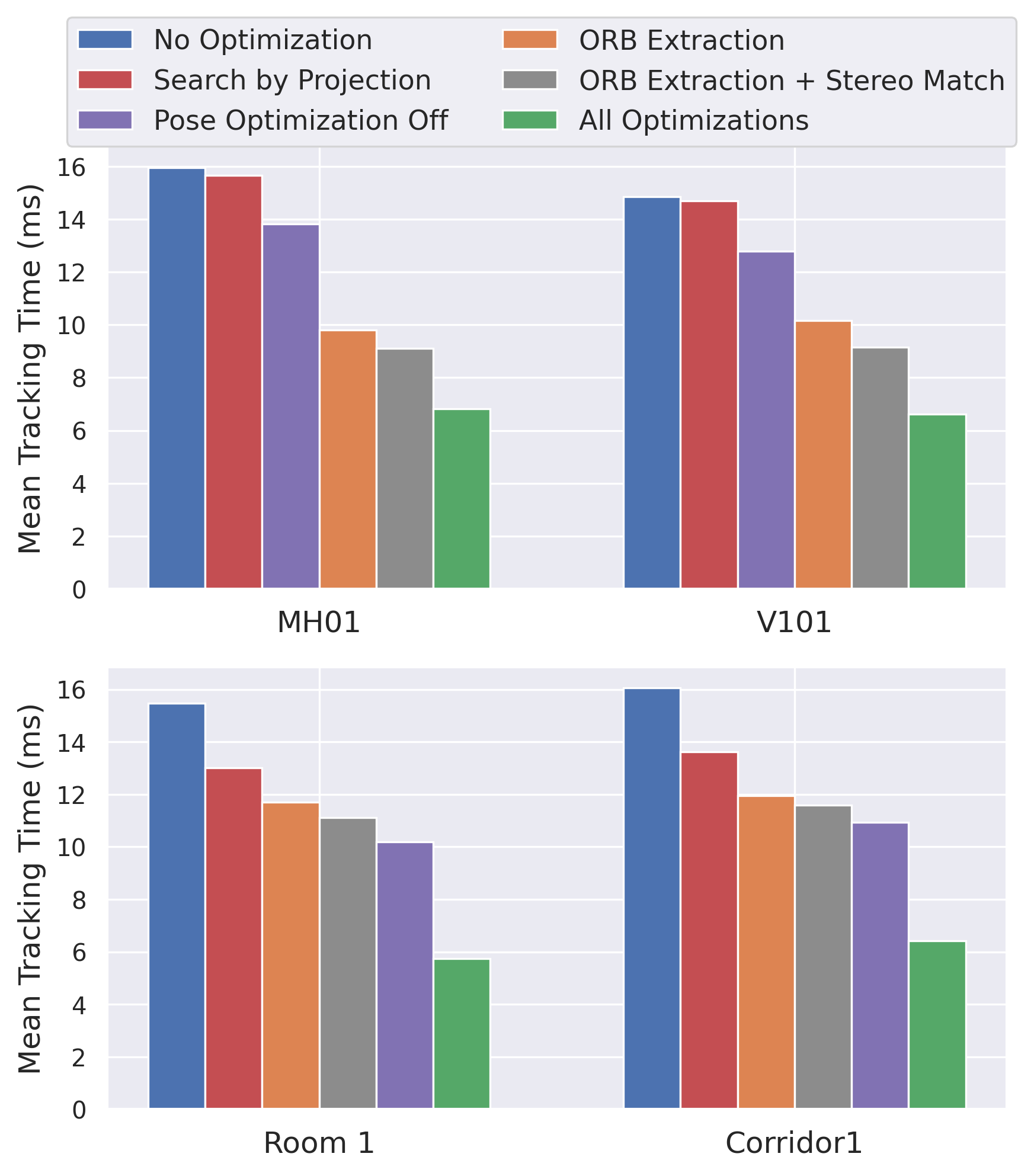}
    \caption{Average tracking time with different optimizations enabled \parsaa{in desktop setting}.}
    \label{fig:performance_comparison}
\end{figure}

\renewcommand{\arraystretch}{1.8}
\begin{table*}[ht]
\vspace{10pt}
\centering
\resizebox{1\linewidth}{!}{
\setlength{\tabcolsep}{4pt}
\begin{tabular}{|c|
*{2}{>{\centering\arraybackslash}p{1.45cm}|}p{0.7cm}|
*{2}{>{\centering\arraybackslash}p{1.45cm}|}p{0.7cm}|
*{2}{>{\centering\arraybackslash}p{1.45cm}|}p{0.7cm}|
*{2}{>{\centering\arraybackslash}p{1.45cm}|}p{0.7cm}|
*{2}{>{\centering\arraybackslash}p{0.7cm}|}}

\hline
\multirow{2}{*}{\textbf{Sequence}} &
\multicolumn{2}{c|}{\textbf{Tracking Time (ms)}} & \multicolumn{1}{c|}{\textbf{Speed}} &
\multicolumn{2}{c|}{\textbf{ORB Extraction (ms)}} & \multicolumn{1}{c|}{\textbf{Speed}} &
\multicolumn{2}{c|}{\textbf{Stereo Match (ms)}} & \multicolumn{1}{c|}{\textbf{Speed}} &
\multicolumn{2}{c|}{\textbf{Track Local Map (ms)}} & \multicolumn{1}{c|}{\textbf{Speed}} & 
\multicolumn{2}{c|}{\textbf{ATE (m)}} \\
\cline{2-3} \cline{5-6} \cline{8-9} \cline{11-12} \cline{14-15}
 & \textbf{Original} & \textbf{FastTrack} & \multicolumn{1}{c|}{\textbf{Up}} 
 & \textbf{Original} & \textbf{FastTrack} & \multicolumn{1}{c|}{\textbf{Up}} 
 & \textbf{Original} & \textbf{FastTrack} & \multicolumn{1}{c|}{\textbf{Up}} 
 & \textbf{Original} & \textbf{FastTrack} & \multicolumn{1}{c|}{\textbf{Up}}
 & \textbf{Org} & \textbf{FT} \\

\hline
\textbf{Average} 
& $14.08$ & $5.47$ & \textcolor{Green}{\textbf{\hspace{+0.04cm}2.57x}} 
& $5.38$ & $1.88$ & \textcolor{Green}{\textbf{\hspace{+0.04cm}2.85x}} 
& $1.10$ & $0.42$ & \textcolor{Green}{\textbf{\hspace{+0.04cm}2.60x}} 
& $6.35$ & $2.11$ & \textcolor{Green}{\textbf{\hspace{+0.04cm}3.00x}} 
& 0.020 & 0.025 \\
\hline
\textbf{MH01} 
& $15.52\pm 2.60$ & $6.41\pm 1.57$ & \textcolor{Green}{\textbf{\hspace{+0.04cm}2.42x}} 
& $7.76\pm 1.14$ & $2.43\pm 0.25$ & \textcolor{Green}{\textbf{\hspace{+0.04cm}3.19x}} 
& $1.29\pm 0.32$ & $0.36\pm 0.04$ & \textcolor{Green}{\textbf{\hspace{+0.04cm}3.58x}} 
& $4.76\pm 1.61$ & $2.14\pm 1.12$ & \textcolor{Green}{\textbf{\hspace{+0.04cm}2.22x}} 
& 0.036 & 0.043 \\
\hline
\textbf{MH02} 
& $14.96\pm 2.23$ & $6.14\pm 1.44$ & \textcolor{Green}{\textbf{\hspace{+0.04cm}2.43x}} 
& $7.70\pm 1.15$ & $2.41\pm 0.27$ & \textcolor{Green}{\textbf{\hspace{+0.04cm}3.19x}} 
& $1.28\pm 0.33$ & $0.36\pm 0.04$ & \textcolor{Green}{\textbf{\hspace{+0.04cm}3.55x}} 
& $4.36\pm 1.40$ & $1.96\pm 1.05$ & \textcolor{Green}{\textbf{\hspace{+0.04cm}2.22x}} 
& 0.030 & 0.027 \\
\hline
\textbf{MH03} 
& $15.40\pm 2.27$ & $5.97\pm 1.12$ & \textcolor{Green}{\textbf{\hspace{+0.04cm}2.57x}} 
& $7.51\pm 1.41$ & $2.37\pm 0.30$ & \textcolor{Green}{\textbf{\hspace{+0.04cm}3.16x}} 
& $1.38\pm 0.35$ & $0.36\pm 0.04$ & \textcolor{Green}{\textbf{\hspace{+0.04cm}3.83x}} 
& $4.73\pm 1.34$ & $1.83\pm 0.74$ & \textcolor{Green}{\textbf{\hspace{+0.04cm}2.58x}} 
& 0.030 & 0.034 \\
\hline
\textbf{MH04} 
& $14.09\pm 2.96$ & $5.80\pm 1.38$ & \textcolor{Green}{\textbf{\hspace{+0.04cm}2.42x}} 
& $6.69\pm 1.18$ & $2.22\pm 0.28$ & \textcolor{Green}{\textbf{\hspace{+0.04cm}3.01x}} 
& $1.47\pm 0.36$ & $0.37\pm 0.05$ & \textcolor{Green}{\textbf{\hspace{+0.04cm}3.97x}} 
& $4.32\pm 1.87$ & $1.85\pm 1.00$ & \textcolor{Green}{\textbf{\hspace{+0.04cm}2.33x}} 
& 0.042 & 0.045 \\
\hline
\textbf{MH05} 
& $14.13\pm 2.78$ & $5.67\pm 1.37$ & \textcolor{Green}{\textbf{\hspace{+0.04cm}2.49x}} 
& $6.71\pm 1.20$ & $2.24\pm 0.28$ & \textcolor{Green}{\textbf{\hspace{+0.04cm}2.99x}} 
& $1.47\pm 0.34$ & $0.37\pm 0.05$ & \textcolor{Green}{\textbf{\hspace{+0.04cm}3.97x}} 
& $4.45\pm 1.67$ & $1.77\pm 0.96$ & \textcolor{Green}{\textbf{\hspace{+0.04cm}2.51x}} 
& 0.055 & 0.075 \\
\hline
\textbf{V101} 
& $14.64\pm 2.01$ & $6.26\pm 1.21$ & \textcolor{Green}{\textbf{\hspace{+0.04cm}2.33x}} 
& $6.71\pm 1.04$ & $2.23\pm 0.26$ & \textcolor{Green}{\textbf{\hspace{+0.04cm}3.00x}} 
& $1.47\pm 0.33$ & $0.37\pm 0.06$ & \textcolor{Green}{\textbf{\hspace{+0.04cm}3.97x}} 
& $4.94\pm 1.32$ & $2.22\pm 0.85$ & \textcolor{Green}{\textbf{\hspace{+0.04cm}2.22x}}
& 0.038 & 0.038\\
\hline
\textbf{V102} 
& $13.99\pm 2.21$ & $5.45\pm 1.00$ & \textcolor{Green}{\textbf{\hspace{+0.04cm}2.56x}} 
& $6.31\pm 0.97$ & $2.17\pm 0.26$ & \textcolor{Green}{\textbf{\hspace{+0.04cm}2.90x}} 
& $1.47\pm 0.43$ & $0.37\pm 0.05$ & \textcolor{Green}{\textbf{\hspace{+0.04cm}3.97x}} 
& $4.71\pm 1.56$ & $1.63\pm 0.66$ & \textcolor{Green}{\textbf{\hspace{+0.04cm}2.88x}}
& 0.018 & 0.020\\
\hline
\textbf{V103} 
& $12.92\pm 2.23$ & $5.16\pm 0.9$ & \textcolor{Green}{\textbf{2.50x}} 
& $6.01\pm 1.04$ & $2.06\pm 0.27$ & \textcolor{Green}{\textbf{2.91x}} 
& $1.38\pm 0.40$ & $0.36\pm 0.05$ & \textcolor{Green}{\textbf{3.83x}} 
& $4.11\pm 1.41$ & $1.52\pm 0.56$ & \textcolor{Green}{\textbf{2.70x}}
& 0.024 & 0.026\\
\hline
\textbf{Room1} 
& $15.39\pm 1.75$ & $5.56\pm 0.80$ & \textcolor{Green}{\textbf{\hspace{+0.04cm}2.76x}} 
& $4.71\pm 0.56$ & $1.79\pm 0.15$ & \textcolor{Green}{\textbf{\hspace{+0.04cm}2.63x}} 
& $0.98\pm 0.08$ & $0.54\pm 0.03$ & \textcolor{Green}{\textbf{1.81x}} 
& $8.60\pm 1.63$ & $2.34\pm 0.60$ & \textcolor{Green}{\textbf{3.67x}}
& 0.011 & 0.011 \\
\hline
\textbf{Room2} 
& $14.66\pm 1.76$ & $5.35\pm 0.74$ & \textcolor{Green}{\textbf{\hspace{+0.04cm}2.74x}} 
& $4.68\pm 0.56$ & $1.78\pm 0.15$ & \textcolor{Green}{\textbf{\hspace{+0.04cm}2.62x}} 
& $0.96\pm 0.09$ & $0.52\pm 0.03$ & \textcolor{Green}{\textbf{\hspace{+0.04cm}1.84x}} 
& $7.98\pm 1.46$ & $2.19\pm 0.53$ & \textcolor{Green}{\textbf{\hspace{+0.04cm}3.64x}} 
& 0.010 & 0.011 \\
\hline
\textbf{Room3} 
& $14.89\pm 1.74$ & $5.40\pm 0.76$ & \textcolor{Green}{\textbf{\hspace{+0.04cm}2.75x}} 
& $4.73\pm 0.60$ & $1.79\pm 0.16$ & \textcolor{Green}{\textbf{\hspace{+0.04cm}2.64x}} 
& $0.97\pm 0.09$ & $0.53\pm 0.03$ & \textcolor{Green}{\textbf{\hspace{+0.04cm}1.83x}} 
& $8.11\pm 1.52$ & $2.19\pm 0.57$ & \textcolor{Green}{\textbf{\hspace{+0.04cm}3.70x}} 
& 0.008 & 0.009 \\
\hline
\textbf{Room4} 
& $15.17\pm 2.03$ & $5.41\pm 0.73$ & \textcolor{Green}{\textbf{\hspace{+0.04cm}2.80x}} 
& $4.60\pm 0.57$ & $1.77\pm 0.15$ & \textcolor{Green}{\textbf{\hspace{+0.04cm}2.59x}} 
& $0.96\pm 0.10$ & $0.53\pm 0.03$ & \textcolor{Green}{\textbf{\hspace{+0.04cm}1.81x}} 
& $8.44\pm 1.87$ & $2.24\pm 0.60$ & \textcolor{Green}{\textbf{\hspace{+0.04cm}3.76x}} 
& 0.009 & 0.008\\
\hline
\textbf{Room5} 
& $15.23\pm 1.58$ & $5.50\pm 0.70$ & \textcolor{Green}{\textbf{\hspace{+0.04cm}2.76x}} 
& $4.62\pm 0.57$ & $1.77\pm 0.15$ & \textcolor{Green}{\textbf{\hspace{+0.04cm}2.61x}} 
& $0.98\pm 0.09$ & $0.53\pm 0.03$ & \textcolor{Green}{\textbf{\hspace{+0.04cm}1.84x}} 
& $8.57\pm 1.41$ & $2.29\pm 0.52$ & \textcolor{Green}{\textbf{\hspace{+0.04cm}3.74x}} 
& 0.010 & 0.010 \\
\hline
\textbf{Room6} 
& $15.57\pm 1.75$ & $5.51\pm 0.69$ & \textcolor{Green}{\textbf{\hspace{+0.04cm}2.82x}} 
& $4.59\pm 0.55$ & $1.76\pm 0.14$ & \textcolor{Green}{\textbf{\hspace{+0.04cm}2.60x}} 
& $0.96\pm 0.09$ & $0.52\pm 0.03$ & \textcolor{Green}{\textbf{\hspace{+0.04cm}1.84x}} 
& $9.02\pm 1.60$ & $2.37\pm 0.56$ & \textcolor{Green}{\textbf{\hspace{+0.04cm}3.80x}} 
& 0.005 & 0.006 \\
\hline
\textbf{Corridor1} 
& $15.71\pm 3.08$ & $6.33\pm 2.02$ & \textcolor{Green}{\textbf{\hspace{+0.04cm}2.48x}} 
& $4.53\pm 0.58$ & $1.72\pm 0.17$ & \textcolor{Green}{\textbf{\hspace{+0.04cm}2.63x}} 
& $0.96\pm 0.10$ & $0.52\pm 0.04$ & \textcolor{Green}{\textbf{\hspace{+0.04cm}1.84x}} 
& $9.11\pm 2.84$ & $3.14\pm 1.86$ & \textcolor{Green}{\textbf{\hspace{+0.04cm}2.90x}} 
& - & 0.002* \\
\hline
\textbf{Corridor2} 
& $15.50\pm 3.45$ & $6.23\pm 2.10$ & \textcolor{Green}{\textbf{\hspace{+0.04cm}2.48x}} 
& $4.55\pm 0.58$ & $1.72\pm 0.16$ & \textcolor{Green}{\textbf{\hspace{+0.04cm}2.64x}} 
& $0.96\pm 0.10$ & $0.52\pm 0.03$ & \textcolor{Green}{\textbf{\hspace{+0.04cm}1.84x}} 
& $9.08\pm 3.16$ & $3.14\pm 1.95$ & \textcolor{Green}{\textbf{\hspace{+0.04cm}2.89x}} 
& - & 0.002* \\
\hline
\textbf{Corridor3} 
& $15.67\pm 2.90$ & $6.47\pm 2.29$ & \textcolor{Green}{\textbf{\hspace{+0.04cm}2.42x}} 
& $4.53\pm 0.56$ & $1.72\pm 0.16$ & \textcolor{Green}{\textbf{\hspace{+0.04cm}2.63x}} 
& $0.96\pm 0.10$ & $0.52\pm 0.03$ & \textcolor{Green}{\textbf{\hspace{+0.04cm}1.84x}} 
& $9.02\pm 2.70$ & $3.28\pm 2.14$ & \textcolor{Green}{\textbf{\hspace{+0.04cm}2.75x}} 
& - & 0.003* \\
\hline
\end{tabular}
}
\caption{Comparison of execution times and their standard deviations, along with ATEs, between \SYS{} and ORB-SLAM3, in our desktop machine. For corridor sequences, the Relative Pose Error (RPE) is reported instead of ATE. See the trajectory section for more details.}
\label{tab:overall_table_std}
\end{table*}
\renewcommand{\arraystretch}{1}

\renewcommand{\arraystretch}{1.8}
\begin{table*}[ht]
\vspace{10pt}
\centering
\resizebox{1\linewidth}{!}{
\setlength{\tabcolsep}{4pt}
\begin{tabular}{|c|
*{2}{>{\centering\arraybackslash}p{1.45cm}|}p{0.7cm}|
*{2}{>{\centering\arraybackslash}p{1.45cm}|}p{0.7cm}|
*{2}{>{\centering\arraybackslash}p{1.45cm}|}p{0.7cm}|
*{2}{>{\centering\arraybackslash}p{1.45cm}|}p{0.7cm}|
*{2}{>{\centering\arraybackslash}p{0.7cm}|}}

\hline
\multirow{2}{*}{\textbf{Sequence}} &
\multicolumn{2}{c|}{\textbf{Tracking Time (ms)}} & \multicolumn{1}{c|}{\textbf{Speed}} &
\multicolumn{2}{c|}{\textbf{ORB Extraction (ms)}} & \multicolumn{1}{c|}{\textbf{Speed}} &
\multicolumn{2}{c|}{\textbf{Stereo Match (ms)}} & \multicolumn{1}{c|}{\textbf{Speed}} &
\multicolumn{2}{c|}{\textbf{Track Local Map (ms)}} & \multicolumn{1}{c|}{\textbf{Speed}} & 
\multicolumn{2}{c|}{\textbf{ATE (m)}} \\
\cline{2-3} \cline{5-6} \cline{8-9} \cline{11-12} \cline{14-15}
 & \textbf{Original} & \textbf{FastTrack} & \multicolumn{1}{c|}{\textbf{Up}} 
 & \textbf{Original} & \textbf{FastTrack} & \multicolumn{1}{c|}{\textbf{Up}} 
 & \textbf{Original} & \textbf{FastTrack} & \multicolumn{1}{c|}{\textbf{Up}} 
 & \textbf{Original} & \textbf{FastTrack} & \multicolumn{1}{c|}{\textbf{Up}}
 & \textbf{Org} & \textbf{FT} \\

\hline
\textbf{Average} 
& $77.49$ & $29.39$ & \textcolor{Green}{\textbf{\hspace{+0.04cm}2.63x}} 
& $28.29$ & $14.08$ & \textcolor{Green}{\textbf{\hspace{+0.04cm}2.00x}} 
& $6.05$ & $2.04$ & \textcolor{Green}{\textbf{\hspace{+0.04cm}2.95x}} 
& $35.01$ & $11.47$ & \textcolor{Green}{\textbf{\hspace{+0.04cm}3.05x}} 
& 0.024 & 0.022 \\
\hline
\textbf{MH01} 
& $84.58\pm 17.3$ & $41.26\pm 8.99$ & \textcolor{Green}{\textbf{\hspace{+0.04cm}2.04x}} 
& $38.83\pm 7.06$ & $18.11\pm 2.36$ & \textcolor{Green}{\textbf{\hspace{+0.04cm}2.14x}} 
& $8.59\pm 2.26$ & $2.54\pm 0.32$ & \textcolor{Green}{\textbf{\hspace{+0.04cm}3.38x}} 
& $28.12\pm 9.31$ & $12.65\pm 5.92$ & \textcolor{Green}{\textbf{\hspace{+0.04cm}2.22x}} 
& 0.038 & 0.036 \\
\hline
\textbf{MH02} 
& $75.70\pm 10.7$ & $40.04\pm 8.38$ & \textcolor{Green}{\textbf{\hspace{+0.04cm}1.89x}} 
& $33.25\pm 3.78$ & $17.98\pm 1.61$ & \textcolor{Green}{\textbf{\hspace{+0.04cm}1.84x}} 
& $8.40\pm 1.87$ & $2.53\pm 0.31$ & \textcolor{Green}{\textbf{\hspace{+0.04cm}3.32x}} 
& $25.52\pm 7.53$ & $11.82\pm 5.93$ & \textcolor{Green}{\textbf{\hspace{+0.04cm}2.15x}} 
& 0.033 & 0.028 \\
\hline
\textbf{Room1} 
& $79.57\pm 10.2$ & $29.33\pm 3.41$ & \textcolor{Green}{\textbf{\hspace{+0.04cm}2.71x}} 
& $24.90\pm 2.91$ & $12.53\pm 0.89$ & \textcolor{Green}{\textbf{\hspace{+0.04cm}1.92x}} 
& $5.74\pm 1.18$ & $1.86\pm 0.14$ & \textcolor{Green}{\textbf{3.08x}} 
& $43.52\pm 6.65$ & $10.09\pm 2.12$ & \textcolor{Green}{\textbf{4.31x}}
& 0.008 & 0.011 \\
\hline
\textbf{Room2} 
& $75.37\pm 9.58$ & $29.19\pm 3.12$ & \textcolor{Green}{\textbf{\hspace{+0.04cm}2.58x}} 
& $24.62\pm 2.46$ & $12.49\pm 0.90$ & \textcolor{Green}{\textbf{\hspace{+0.04cm}1.97x}} 
& $5.41\pm 0.91$ & $1.81\pm 0.15$ & \textcolor{Green}{\textbf{\hspace{+0.04cm}2.98x}} 
& $40.04\pm 5.96$ & $10.09\pm 2.06$ & \textcolor{Green}{\textbf{\hspace{+0.04cm}3.96x}} 
& 0.022 & 0.011 \\
\hline
\textbf{Corridor1} 
& $75.41\pm 11.4$ & $30.38\pm 6.16$ & \textcolor{Green}{\textbf{\hspace{+0.04cm}2.48x}} 
& $23.98\pm 2.49$ & $11.71\pm 1.20$ & \textcolor{Green}{\textbf{\hspace{+0.04cm}2.04x}} 
& $4.12\pm 0.69$ & $1.78\pm 0.18$ & \textcolor{Green}{\textbf{2.31x}} 
& $41.60\pm 9.69$ & $11.73\pm 5.21$ & \textcolor{Green}{\textbf{3.54x}}
& - & 0.002* \\
\hline
\textbf{Corridor2} 
& $74.31\pm 12.7$ & $30.72\pm 6.88$ & \textcolor{Green}{\textbf{\hspace{+0.04cm}2.41x}} 
& $24.20\pm 2.70$ & $11.70\pm 1.35$ & \textcolor{Green}{\textbf{\hspace{+0.04cm}2.06x}} 
& $4.07\pm 0.77$ & $1.76\pm 0.17$ & \textcolor{Green}{\textbf{\hspace{+0.04cm}2.31x}} 
& $41.28\pm 9.52$ & $12.48\pm 5.73$ & \textcolor{Green}{\textbf{\hspace{+0.04cm}3.30x}} 
& - & 0.002* \\
\hline
\end{tabular}
}
\caption{Comparison of execution times and their standard deviations, along with ATEs, between \SYS{} and ORB-SLAM3 in Jetson Xavier NX. For corridor sequences, RPEs are reported instead of ATEs. See the trajectory section for more details.}
\label{tab:overall_table_std_jetson}
\end{table*}
\renewcommand{\arraystretch}{1}

\begin{table}[t]
    \vspace{10pt}
    \centering
    \footnotesize
    \setlength{\tabcolsep}{4pt}
    \renewcommand{\arraystretch}{1.2} 
    \resizebox{\linewidth}{!}{
        \begin{tabular}{|c|c|c|c|c|}
        \hline
        \multirow{2}{*}{Sequence} & \multicolumn{2}{c|}{ORB-SLAM3} & \multicolumn{2}{c|}{FastTrack} \\
        \cline{2-5}
        & Avg Loss & Zero-Drop & Avg Loss & Zero-Drop \\
        \hline
        MH04       & 189.8 & 0/5    & 43.0  & 4/5 \\
        \hline
        MH05       & 126.6 & 1/5   & 66.7  & 4/5 \\
        \hline
        Room1      & 50.2  & 2/5   & 0.0   & \textbf{5/5} \\
        \hline
        Room2      & 43.2  & 2/5   & 0.0   & \textbf{5/5} \\
        \hline
        \end{tabular}
    }
    \caption{Frame loss comparison between ORB-SLAM3 and FastTrack (desktop)}
    \label{tab:frame-lost-desktop}
\end{table}


\begin{table}[t]
    \vspace{10pt}
    \centering
    \footnotesize
    \setlength{\tabcolsep}{4pt}
  \renewcommand{\arraystretch}{1.2} 
  \resizebox{\linewidth}{!}{
  \begin{tabular}{|c|c|c|c|c|}
    \hline
    \multirow{2}{*}{Sequence} & \multicolumn{2}{c|}{ORB-SLAM3} & \multicolumn{2}{c|}{FastTrack} \\
    \cline{2-5}
    & Avg Loss & Zero-Drop & Avg Loss & Zero-Drop \\
    \hline
    MH01 & 50.2 & 2/5 & 0.0 & \textbf{5/5} \\
    \hline
    MH02 & 69.4 & 3/5 & 0.0 & \textbf{5/5} \\
    \hline
    Room1 & 66.40 & 0/5 & 33.20 & 2/5 \\
    \hline
    Room2 & 58.80 & 0/5 & 10.60 & 4/5 \\
    \hline
  \end{tabular}
  }
  \caption{Frame loss comparison between ORB-SLAM3 and FastTrack (Jetson)}
  \label{tab:frame-lost-jetson}
\end{table}

\subsection{Experimental Setup}
We compare the performance of different tasks in tracking between ORB-SLAM3 and \SYS{}. \kimia{The experiments are run on a desktop machine and an NVIDIA Jetson Xavier NX board, described in \autoref{tab:specs}.} We run a mix of sequences from EuRoC and TUM-VI datasets using the stereo-inertial configuration. We report the average results over five runs for each sequence in \parsaa{all of the following sections}. \parsa{}

\subsection{Overall Performance}
\autoref{tab:overall_table_std} and \autoref{tab:overall_table_std_jetson} show the performance of \SYS{} compared to ORB-SLAM3 \kimia{in our desktop and Jetson settings, respectively}. We compare tracking time and ATE (RMSE) in both systems along with the impact of each optimization individually on the corresponding component's timing. \kimia {In the timings reported for FastTrack, GPU data transfer times are also included.} 

Our results demonstrate a significant improvement in tracking times, achieving up to 2.8$\times$ faster performance \kimia{on desktop, and up to 2.7$\times$ speed up on Xavier NX}. \kimia{On desktop, \SYS{} achieves an average tracking time of 5.5 ms per frame, corresponding to an approximate processing rate of 182 FPS}. \kimia{On the Jetson device, FastTrack achieves an average tracking time of 29.4 ms per frame, corresponding to a processing rate of 34 FPS.} 
The results also demonstrate minimal change in the ATEs, showing that our optimizations do not affect the system's accuracy.

Furthermore, our implementation significantly tightens the standard deviation of the average times across all components. \kimia{By limiting the number of operations on the CPU, and thereby reducing the burden of multitasking and moving tasks to the GPU, we achieve a 45\% reduction in overall tracking time variance.}
This reduction in variance helps prevent frame drops by reducing the maximum tracking time, ensuring more consistent and timely frame handling. To illustrate this point further, \autoref{fig:room1_tracking_time} compares tracking times \kimia{on the desktop} for Room1 between \SYS{} and ORB-SLAM3. It serves as a representative example, with similar outcomes observed across other \kimia{experiments}. This figure highlights the reduction of average tracking time and the tightening of the standard deviation. Moreover, the reduction of maximum tracking time from 20ms to 8ms highlights the system's robustness in handling diverse frames. \autoref{tab:frame-lost-desktop} and \autoref{tab:frame-lost-jetson} show frame loss comparison between ORB-SLAM3 and FastTrack on our desktop machine and Jetson device, respectively. We report the average number of frames lost per sequence, as well as the ratio of experiments with zero frame drops. As shown in the tables, FastTrack consistently results in fewer frame drops and completely eliminates them in some sequences, such as Room1 and Room2 on the desktop, and MH01 and MH02 on the Jetson.

\autoref{fig:performance_comparison} shows the mean tracking time of the system \kimia{on our desktop} running with \parsa{individual optimization strategies discussed in \autoref{sec:design}}, with the last bar in each group representing all optimizations enabled. 
We evaluate each optimization individually by isolating each from all others to compare their performance separately, except for the stereo matching kernel. We only run the stereo matching kernel with the ORB Extraction kernel on to avoid high CPU-to-GPU data transfer overhead between them as explained earlier.

As shown in \autoref{fig:performance_comparison}, ORB extraction has the most impact on system performance in EuRoC sequences. However, since local map tracking is more time-consuming in the TUM-VI 
sequences, the impact of optimized components on the speedup shifts, and disabling Pose Optimization becomes the most significant factor in speeding up tracking time.

\subsection{Performance of Individual Design Strategies}
\subsubsection{Stereo Matching} 
\kimia{Both} \autoref{tab:overall_table_std} \kimia{and \autoref{tab:overall_table_std_jetson}} show a consistent speedup of stereo feature matching in both EuRoC and TUM-VI datasets. \kimia{In our desktop experiments,} we observe a speedup of up to 3.97$\times$ in EuRoC, and up to 1.84$\times$ in TUM-VI. \kimia{On Jetson, we observe a speedup of up to 3.38$\times$ in EuRoC, and up to 3.08$\times$ in TUM-VI}. The variation in speedup can be attributed to the different design strategies used for stereo matching in pinhole versus fisheye settings, as detailed in \autoref{sec:design}. We achieve better speedups in EuRoC compared to TUM-VI since the logic is more parallelizable and we utilize more GPU threads for calculations in the EuRoC dataset, which uses a pinhole camera.

\subsubsection{Track Local Map} 
As shown in \autoref{tab:overall_table_std} \kimia{and \autoref{tab:overall_table_std_jetson}}, our optimization achieves up to 3.8$\times$ speedup in tracking the local map \kimia{on desktop, and up to 4.31$\times$ on Jetson.} The speedup is more significant in TUM-VI compared to EuRoC. This difference again arises from the different camera types used in these datasets. In the pinhole setting, only keypoints from the left image are used for tracking the local map. However, in the fisheye setting, keypoints from both left and right images are involved. This doubles the execution time and the number of iterations needed, making GPU optimizations more impactful.

The performance improvement in tracking the local map is due to executing Search by Projection on the GPU and disabling the Pose Optimization component. The following sections will individually discuss the impact of these changes \parsaa{in the desktop setting, with similar results observed in Jetson}.

\noindent\textbf{Search By Projection Kernel:}
In \autoref{fig:performance_comparison}, the second bar of both plots represents the mean tracking time when Search by Projection is executed on the GPU. As previously discussed, GPU optimization has more impact on Room1 and Corridor1 (TUM-VI) compared to MH01 and V101 (EuRoC). Our findings indicate that using the Search by Projection kernel results in approximately a 1.2$\times$ tracking time speedup for TUM-VI sequences, while the speedup for EuRoC sequences is negligible and is only about 1.01$\times$. 

\noindent\textbf{Disabling Pose Optimization:}
The bars labeled Pose Optimization Off in \autoref{fig:performance_comparison} represent the mean tracking time when pose optimization is disabled inside Track Local Map. Similar to Track Local Map and Search by Projection, disabling pose optimization has a greater impact on tracking time for Room and Corridor sequences compared to Machine Hall and Vicon. Our experiments show that disabling pose optimization leads to approximately a 1.5$\times$ increase in tracking speed for TUM-VI and around a 1.1$\times$ speed increase for EuRoC, while not having much impact on tracking accuracy as mentioned in \autoref{design_pose_optimization}.

\subsection{Trajectory} \label{sec:trajectory}
\begin{figure}[t]
    \centering
    \includegraphics[width=.9\linewidth]{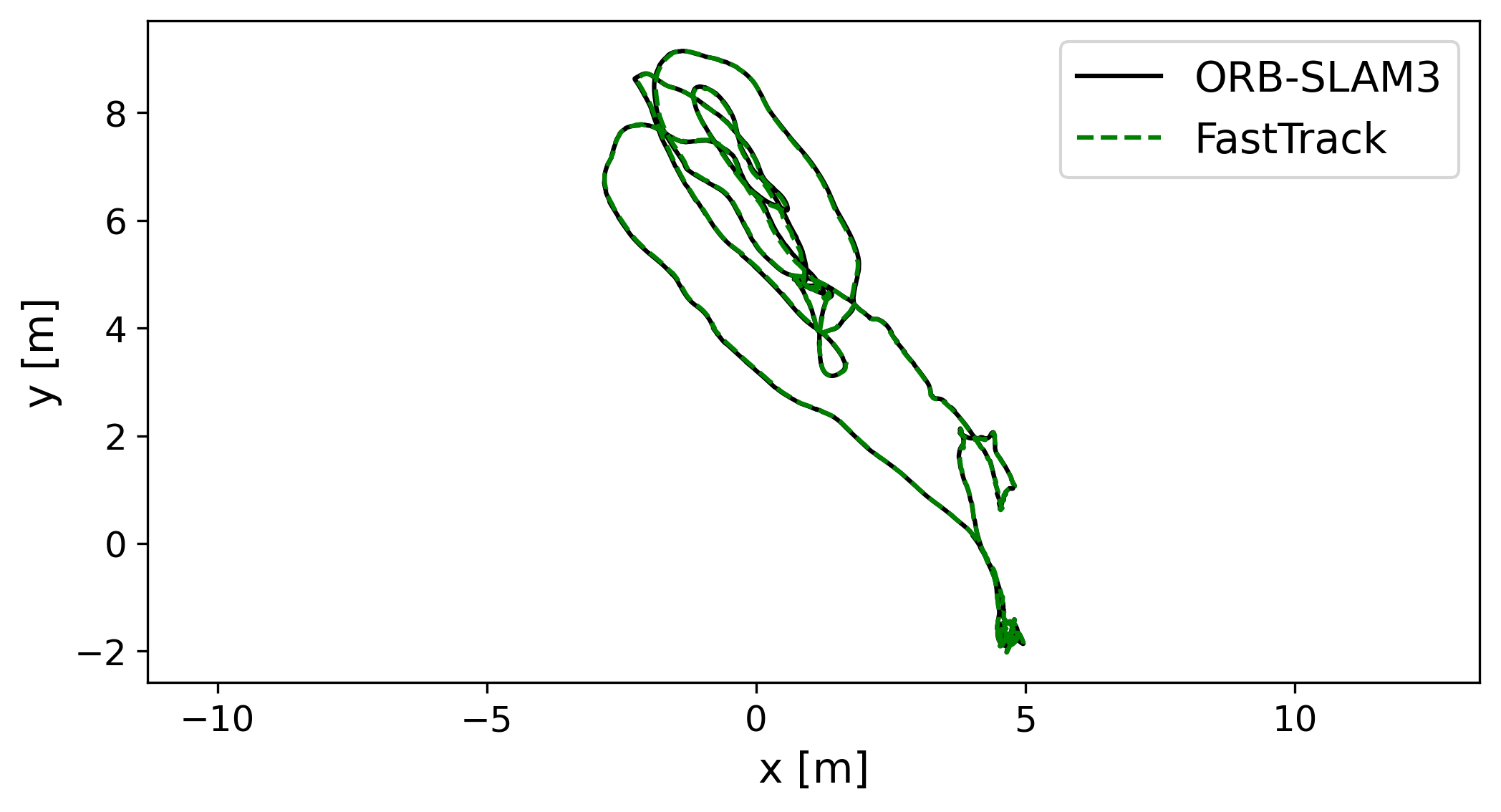}
    \caption{MH01 trajectory comparison in ORB-SLAM3 and \SYS{}}
    \label{fig:trajectory_comparison}
\end{figure}

We compare the trajectory of \SYS{} to that of ORB-SLAM3. \autoref{fig:trajectory_comparison} shows the trajectory results for \kimia{MH01} as an example. As shown, the trajectories are closely aligned, leading to similar ATE values. This trend is consistent across other sequences as well, except for the corridor sequences.
The corridor sequences lack complete ground truth, which is only available at the beginning of each sequence. As a result, proper alignment of the FastTrack trajectory with the ground truth is not possible, leading to invalid ATE values. Additionally, ORB-SLAM3 trajectory cannot be used as a ground truth reference for ATE evaluation, as its results exhibit slight variations across runs due to inherent randomness. To address this limitation, we computed an alternative metric, Relative Pose Error (RPE), which measures the local accuracy of motion between consecutive frames. We compare the motion estimated by FastTrack to that of ORB-SLAM3. As shown in \autoref{tab:overall_table_std} and \autoref{tab:overall_table_std_jetson}, the RPE for all corridor sequences in FastTrack does not exceed a few millimeters, demonstrating that our system performs with accuracy comparable to ORB-SLAM3 in these sequences.

We also observe that \SYS{} achieves a lower ATE compared to ORB-SLAM3 on Jetson in most of the sequences (excluding corridor sequences for the reasons discussed above). This is because ORB-SLAM3 often fails to process new frames in a timely manner in Jetson, resulting in lost frames and lower accuracies. In contrast, \SYS{} processes every frame promptly, resulting in fewer frame losses. For example, ORB-SLAM3 drops 58 frames on average in Room2, while \SYS{} does not drop any frame at all, leading to a lower ATE.


\subsection{Comparative Evaluation}
\label{subsec:comparison}

\autoref{tab:comparison} compares the FPS and ATE of ORB-SLAM3, \SYS{}, and
Jetson-SLAM (with its ORB-SLAM2 backend)~\cite{KumarPark24}, in a desktop
stereo setup. We use the Jetson-SLAM code from the main branch of their GitHub
repository as-is with one exception, where we re-enable the code that processes
each frame in real time based on the timestamps of the data set rather than
as quickly as possible. This is for fairness as disabling real-time processing
produces worse ATE results for Jetson-SLAM, and it is also the default setting
for ORB-SLAM2, ORB-SLAM3, and our system.

As observed in the table, \SYS{} achieves the highest FPS in sequences MH01-MH03,
while Jetson-SLAM has higher FPS for MH04 and MH05 at the expense of increased ATE.
We note that the Jetson-SLAM paper~\cite{KumarPark24} reports MH01-MH03 results
but not MH04 and MH05. Our ATE results for MH01-MH03 align with the reported
results, though we observe lower FPS in our experiments.

\begin{table}
  \centering
  \renewcommand{\arraystretch}{1.2} 
  \begin{tabular}{|c|c|c|c|c|c|c|}
    \hline
    \multicolumn{1}{|c|}{\multirow{2}{*}{Sequence}} & \multicolumn{2}{c|}{ORB-SLAM3} & \multicolumn{2}{c|}{FastTrack} & \multicolumn{2}{c|}{Jetson-SLAM} \\
    \cline{2-7}
    & FPS & ATE & FPS & ATE & FPS & ATE \\
    \hline
    MH01 &74.57 & 0.03 & 135.57 & 0.03 & 92.58 & 0.03 \\
    \hline
    MH02 & 78.51 & 0.02 & 144.29 & 0.02 & 106.03 & 0.06 \\
    \hline
    MH03 & 74.91 & 0.03 & 138.29 & 0.03 & 104.10 & 0.06 \\
    \hline
    MH04 & 75.30 & 0.08 & 130.99 & 0.06 & 160.20 & 0.26 \\
    \hline
    MH05 & 75.72 & 0.05 & 137.00 & 0.07 & 142.06 & 0.18 \\
    \hline
  \end{tabular}
  \caption{FPS and ATE comparison between the systems.}
  \label{tab:comparison}
\end{table}


\section{Conclusion}



In this paper, we introduce \SYS{}, which utilizes GPU resources to address tracking bottlenecks in visual-inertial SLAM. 
We design and implement several strategies to accelerate time-consuming parts of tracking in ORB-SLAM3, including stereo feature matching and \kw{search by projection}. We further boost the performance by integrating feature extraction acceleration, handling data transfers efficiently between different tracking components, and bypassing pose optimization.
Through our experiments using EuRoC and TUM-VI datasets, we show that our design achieves an overall speedup of up to 2.8$\times$ for the tracking process in stereo-inertial mode while maintaining comparable accuracy to the original system.

\addtolength{\textheight}{-12cm}   

\bibliographystyle{IEEETran}
\bibliography{references}

\end{document}